%% file: _main.tex
\input{_constants}
\arxiv 

\pdfoutput=1
\documentclass[10pt,twocolumn,letterpaper]{article}
\input{cvpr_header}

\unless\ifarxiv \myexternaldocument{_supplementary} \fi

\begin{document}

\title{Human-Robot Navigation using Event-based Cameras and Reinforcement Learning}
\author{\authorBlock}
\maketitle

\input{00_abstract}
\input{01_intro}
\input{02_related}

\input{03_preliminaries}

\input{04_method}

\input{05_results}

\input{10_conclusion}

{\small
\bibliographystyle{ieeenat_fullname}
\bibliography{11_references}
}


\end{document}

%% file: _constants.tex
\def\paperID{0000}
\def\confName{CVPR}
\def\confYear{2025}

\def\authorBlock{
    Ignacio Bugueno-Cordova$^{1,3}$ \qquad
    Javier Ruiz-del-Solar$^{1,2}$ \qquad
    Rodrigo Verschae$^{3}$ \\
    $^1$Department of Electrical Engineering, Universidad de Chile \\
    $^2$Advanced Mining Technology Center (AMTC), Universidad de Chile \\
    $^3$Institute of Engineering Sciences, Universidad de O’Higgins\\
    {\tt\small ignacio.bugueno@\{ing.uchile.cl, uoh.cl\} 
    \qquad 
    jruizd@ing.uchile.cl 
    \qquad 
    rodrigo@verschae.org}
}

\newif\ifreview 
\newif\ifarxiv \newcommand{\arxiv}{\arxivtrue}
\newif\ifcamera 
\newif\ifrebuttal 

%% file: cvpr_header.tex
\ifreview \usepackage[review]{cvpr} \fi
\ifarxiv \usepackage[pagenumbers]{cvpr} \fi
\ifrebuttal \usepackage[rebuttal]{cvpr} \fi
\ifcamera \usepackage{cvpr} \fi

\input{_macros}  

\usepackage{xr-hyper}

\makeatletter
\newcommand*{\addFileDependency}[1]{
  \typeout{(#1)}
  \@addtofilelist{#1}
  \IfFileExists{#1}{}{\typeout{No file #1.}}
}

\makeatother
\newcommand*{\myexternaldocument}[1]{
    \externaldocument{#1}
    \addFileDependency{#1.tex}
    \addFileDependency{#1.aux}
}

\definecolor{cvprblue}{rgb}{0.21,0.49,0.74}
\usepackage[pagebackref,breaklinks,colorlinks,allcolors=cvprblue]{hyperref}
\usepackage[capitalize]{cleveref}
\crefname{section}{Sec.}{Secs.}
\crefname{table}{Table}{Tables}
\crefname{figure}{Fig.}{Figs.}

\ifarxiv \crefname{appendix}{App.}{Apps.}
\else \crefname{appendix}{Suppl.}{Suppls.} \fi

%% file: _macros.tex
\usepackage{graphicx}	
\usepackage{amsmath}	
\usepackage{amssymb}	
\usepackage{booktabs}
\usepackage{times}
\usepackage{microtype}
\usepackage{epsfig}
\usepackage{caption}
\usepackage{float}
\usepackage{placeins}
\usepackage{color, colortbl}
\usepackage{stfloats}
\usepackage{tabularx}
\usepackage{xstring}
\usepackage{multirow}
\usepackage{xspace}
\usepackage{url}
\usepackage{subcaption}
\usepackage{xcolor}
\usepackage[hang,flushmargin]{footmisc}
\usepackage{algorithm}
\usepackage{algpseudocode}
\usepackage{enumitem}
\usepackage[accsupp]{axessibility}  

\ifcamera \usepackage[accsupp]{axessibility} \fi

% ...

\ifarxiv  \fi

\newcommand{\R}[1]{{%
    \textbf{%
        \ifstrequal{#1}{1}{\textcolor{red}{R#1}}{%
        \ifstrequal{#1}{2}{\textcolor{blue}{R#1}}{%
        \ifstrequal{#1}{3}{\textcolor{magenta}{R#1}}{%
        \ifstrequal{#1}{4}{\textcolor{teal}{R#1}}{%
                           \textcolor{cyan}{R#1}%
        }}}}%
    }%
}}

%% file: 00_abstract.tex
\begin{abstract}
This work introduces a robot navigation controller that combines event cameras and other sensors with reinforcement learning to enable real-time human-centered navigation and obstacle avoidance. 
Unlike conventional image-based controllers, which operate at fixed rates and suffer from motion blur and latency, this approach leverages the asynchronous nature of event cameras to process visual information over flexible time intervals, enabling adaptive inference and control.
The framework integrates event-based perception, additional range sensing, and policy optimization via Deep Deterministic Policy Gradient, with an initial imitation learning phase to improve sample efficiency. Promising results are achieved in simulated environments, demonstrating robust navigation, pedestrian following, and obstacle avoidance. A demo video is available at the \href{https://ibugueno.github.io/hr-navigation-using-event-cameras-and-rl/}{project website}.

\end{abstract}

%% file: 01_intro.tex
\section{Introduction}
\label{sec:intro}

Reinforcement Learning (RL) addresses several applications from robotic manipulation to planning \cite{singh2022reinforcement,huang2023,garaffa2023}. A key aspect is how an agent learns from environmental interactions, integrating rewards to develop a policy that maximizes cumulative rewards over time.

This work introduces a novel controller for human-centered robot navigation that integrates event-based vision and RL. In contrast to frame-based cameras, event cameras offer microsecond-level resolution, high dynamic range, and minimal motion blur~\cite{Gallego2019-wf}, making them suitable for robot navigation and human tracking.
The proposed framework combines data from event cameras, depth, and LIDAR sensors. It employs a two-stage learning process: an initial imitation learning (IL) phase generates a suboptimal policy, followed by refinement using Deep Deterministic Policy Gradients (DDPG). Pedestrian detection is performed using a YOLOv5s model trained on the PEDRo dataset with SAE representations and a 10~ms event accumulation window.

The main contributions of this work are:
\begin{enumerate*}[label=\roman*)]
    \item the first integration of event-based vision with reinforcement learning for human-robot navigation;
    \item a hybrid IL+RL training pipeline that reduces the learning time of the RL agent in a simulated environment;
    \item a quantitative comparison of PD, IL, and RL controllers for person-following and obstacle avoidance.
\end{enumerate*}

The rest of this paper is organized as follows: 
Section~\ref{sec:related-works} reviews related work; 
Section~\ref{sec:preliminaries} introduces fundamental of event-based cameras and robot kinematics; 
Section~\ref{sec:framework} describes the proposed navigation framework; 
Section~\ref{sec:results} presents the experimental results; 
and Section~\ref{sec:conclusions} concludes the paper and projects this work.

%% file: 02_related.tex
\section{Related Work}
\label{sec:related-works}

\subsection{Robot navigation using reinforcement learning}

Recent advances in Deep Reinforcement Learning (DRL) have driven significant progress in Image-Based Visual Servo (IBVS) for robotics. 
In \cite{shinde2019}, the authors introduce a DRL-IBVS framework leveraging DDPGs to track dynamic objects in unmanned aerial vehicles. This method demonstrated effective object tracking validated in Gazebo simulations by directly mapping errors from the \textit{bounding box} to linear velocity commands. 
In \cite{albekairi2023}, the authors propose a collision-free IBVS approach with image plane trajectory planning to improve obstacle avoidance and maintain target visibility, focusing on the stability of system dynamics for mobile robots.
Also, \cite{jin2022} addressed visibility constraints by introducing a DRL-based IBVS controller that adapts control gains, ensuring continuous visibility of a static object. Similarly, \cite{sampedro2018} developed a DDPG-based RL-IBVS controller for multirotor robots, showing the system's efficiency in object tracking tasks in simulated and real scenarios.
Finally, \cite{fu2023} addressed parameter tuning within the visual servo of robots, proposing a DRL framework to optimize kinematic parameters dynamically in several tasks. 

\subsection{Reinforcement learning for robotics using event cameras}

Event cameras have boosted deep learning applications in robotics by enabling fast response in complex environments. In \cite{arakawa2020}, the authors focused on a precedent by integrating RL with event cameras for the first time in car-like robots, creating image-like features from event streams for tasks such as fast collision avoidance and obstacle tracking.
On the other hand, \cite{souissi2024} advanced the field in unmanned aerial vehicle (UAV) tracking by developing a deep reinforcement learning framework that maps event streams to drone control actions. Leveraging event cameras' high dynamic range and low latency, they achieved effective tracking in challenging conditions, including variable illumination, through domain randomization.
Finally, \cite{hu2022} presented a detection and avoidance method for UAVs by directly processing individual events using a variational event-based autoencoder and a continuous-action RL controller. Their real-time obstacle avoidance system achieved higher success rates in navigation tasks than frame-based methods, highlighting the potential of event cameras in high-speed robotic applications.

%% file: 03_preliminaries.tex
\section{Preliminaries}
\label{sec:preliminaries}

\subsection{Event camera data}

Event cameras respond asynchronously to brightness changes in the logarithmic intensity \( L(\textbf{u}_k, t_k) \doteq \text{log}(I(\textbf{u}_k, t_k)) \) at pixel \( \textbf{u}_k \) and time \( t_k \). An event is triggered at \( \textbf{u}_k = (x_k, y_k)^T \) and \( t_k \) when brightness change surpasses a threshold \( \pm C \), where \( p_k \in \{-1, +1\} \) indicates polarity. Within a time window, the camera outputs an event sequence 
\( \mathcal{E}(t_N) = \{ (\textbf{u}_k, t_k, p_k) \}_{k=1}^{N} \) 
with \( \mu s \) resolution \cite{Gallego2019-wf}.

\subsection{Event data representation}

Extracting useful information from event streams is imperative. Representations include: 
\begin{enumerate*}[label=\roman*)]
\item event frames (2D histograms), 
\item temporal surfaces (2D pixel-wise timestamps), 
\item voxel grids (3D spatio-temporal histograms), 
\item reconstructed frames, among others \cite{Gallego2019-wf}.
\end{enumerate*}
One common method is the Surface of Active Events (SAE) \cite{Benosman2014, Lagorce2017-ka}, dividing events by polarity. For each pixel \((x, y)\), the last event within a fixed interval \(\Delta t \) is scaled to \([0, 255]\), producing positive and negative channels:
\begin{equation}
    \text{SAE}_{\text{neg/pos}}(x, y) = 255 \left( \frac{t_e - t_{\text{init}}}{\Delta t} \right).
\end{equation}
Both channels combine to create a composite image \( I_{SAE}(x, y) \), capturing recent activity.

\subsection{Human-centered social robot}

\begin{figure}[!h]
\centering
    \includegraphics[width=\linewidth]{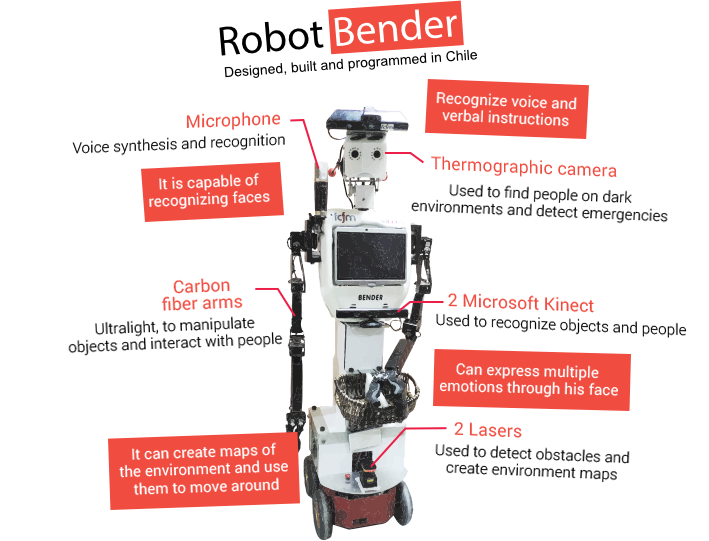}
    \caption{Bender: A General-Purpose Social Robot for Human-Robot Interaction \cite{bender2013}.}
    \label{fig:bender}
\end{figure}

Bender \cite{bender2013} is a humanoid service robot from the University of Chile designed for human-robot interaction (HRI) in domestic environments.
The main idea behind Bender's design was to create an open and flexible research platform that offers human-like communication capabilities and empathy. Bender has an anthropomorphic upper body and a differential drive platform that provides mobility, as shown in Figure~\ref{fig:bender}.

\begin{figure}[!h]
\centering
    \includegraphics[width=0.7\linewidth]{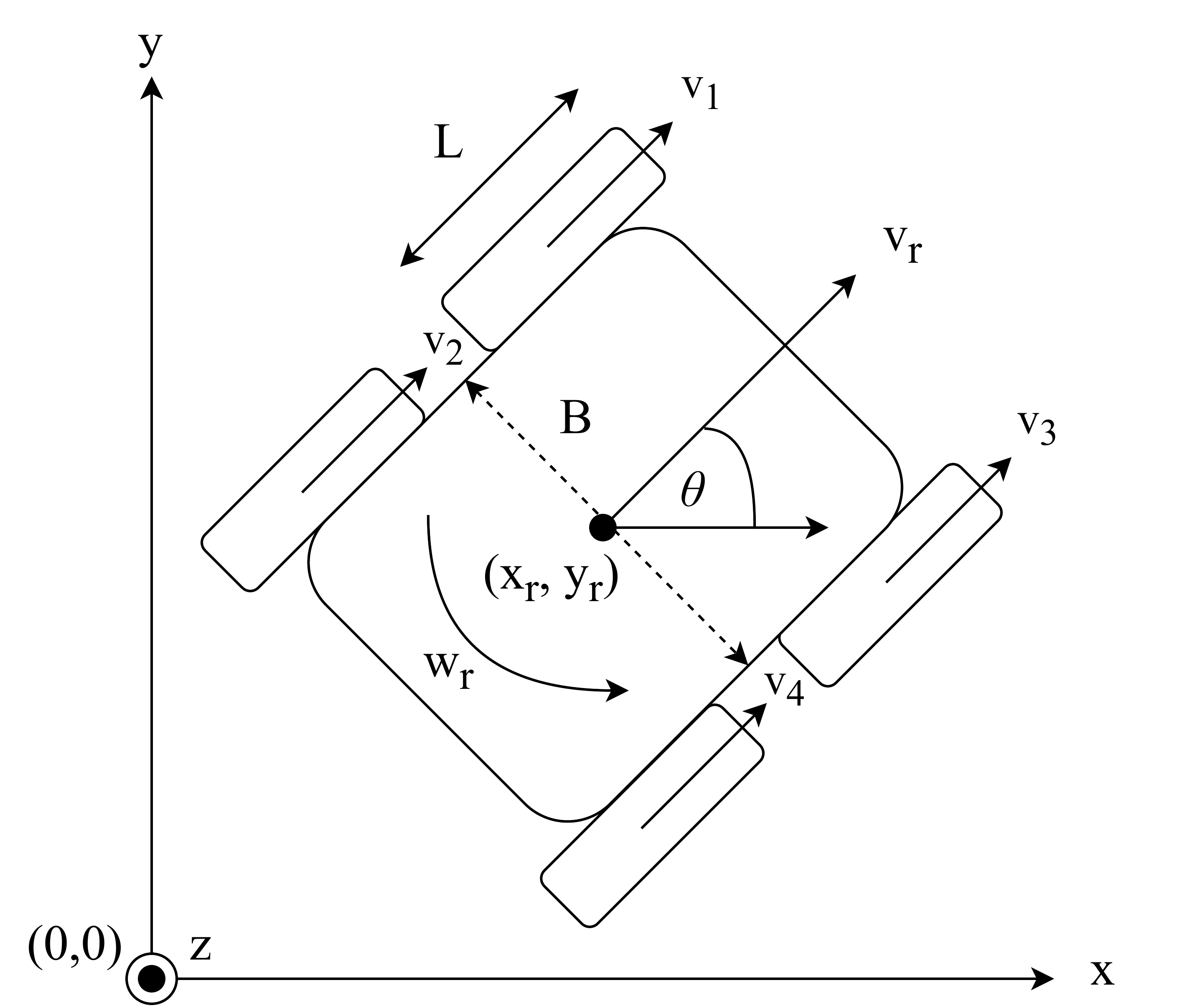}
    \caption{Kinematics of a four-wheel mobile robot.}
    \label{fig:wheel_kinematics}
\end{figure}

Equipped with a Pioneer3AT \cite{pioneer3at}, the robot platform allows a maximum linear acceleration of \( 2 \, [m/s^2] \) and angular acceleration of \( 5.24 \, [rad/s^2] \). For a $0.1[s]$ reaction time, velocity increments are limited at \( 0.2 \, [m/s] \) (linear) and \( 0.5 \, [rad/s] \) (angular) for navigation control design. Figure \ref{fig:wheel_kinematics} provides a schematic of the robotic platform's kinematics. 


In the illustration, \((x_r, y_r)\) correspond to the coordinates of the midpoint between the driving wheels, \( \theta \) is the orientation of the robot, $\{v_i\}_{i=1}^4$ and $\{\omega_i\}_{i=1}^4$ are the linear and angular speeds of the four wheels of the mobile robotic platform, \(B\) is the distance between the wheels, and \(L\) is the diameter of each wheel. It is assumed that the motion of the robot is without slippage and also that $v_1 = v_2$, $v_3 = v_4$, $\omega_1 = \omega_2$, $\omega_3 = \omega_4$. Then, the kinematic model of the robot can be expressed as:
\begin{equation}
\begin{cases}
    \dot{x}_r = v_r \cos \theta, \\
    \dot{y}_r = v_r \sin \theta, \\
    \dot{\theta} = \omega_r,
\end{cases}
\end{equation}
where \(v_r \) is the linear velocity of the robot given by
\(    
    v_r = \frac{L}{4} (\omega_1 + \omega_3),
\)
and \(\omega_r\) is the angular velocity of the robot, expressed by
\(
    \omega_r = \frac{L}{2B} (\omega_3 - \omega_1).
\)
The state of the mobile robot is given by \( q = \left[ x_r, y_r, \theta \right]^T \), and the two control inputs are \( v_r \) and \( \omega_r \).

%% file: 04_method.tex
\section{Proposed Framework}
\label{sec:framework}

\begin{figure}[!h]
\centering
    \includegraphics[width=\linewidth]{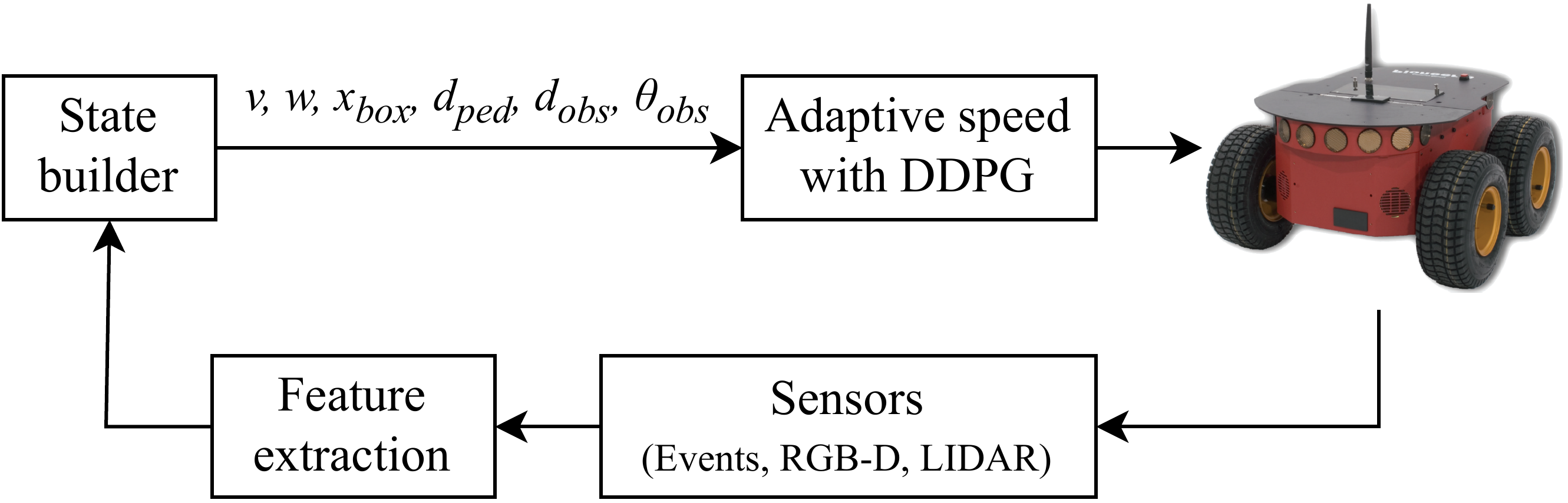}
    \caption{Human-robot navigation framework proposal.}
    \label{fig:proposal}
\end{figure}

This section presents the proposed framework for human-robot navigation using event-based cameras and reinforcement learning. We introduce the pipeline of Figure \ref{fig:proposal}, which includes:
\begin{enumerate*}[label=\roman*)]
\item the sensors employed (event cameras, depth sensors, and obstacle detection),
\item a feature extraction module (in which an event-based pedestrian detector is employed),
\item the system state builder, 
\item and the use of a model-free off-policy reinforcement learning algorithm (DDPG) for the adaptive calculation of the linear and angular velocities of the robotic platform.
\end{enumerate*}

\subsection{Sensors on the mobile robotic platform}

\begin{figure}[!h]
\centering
    \includegraphics[width=0.7\linewidth]{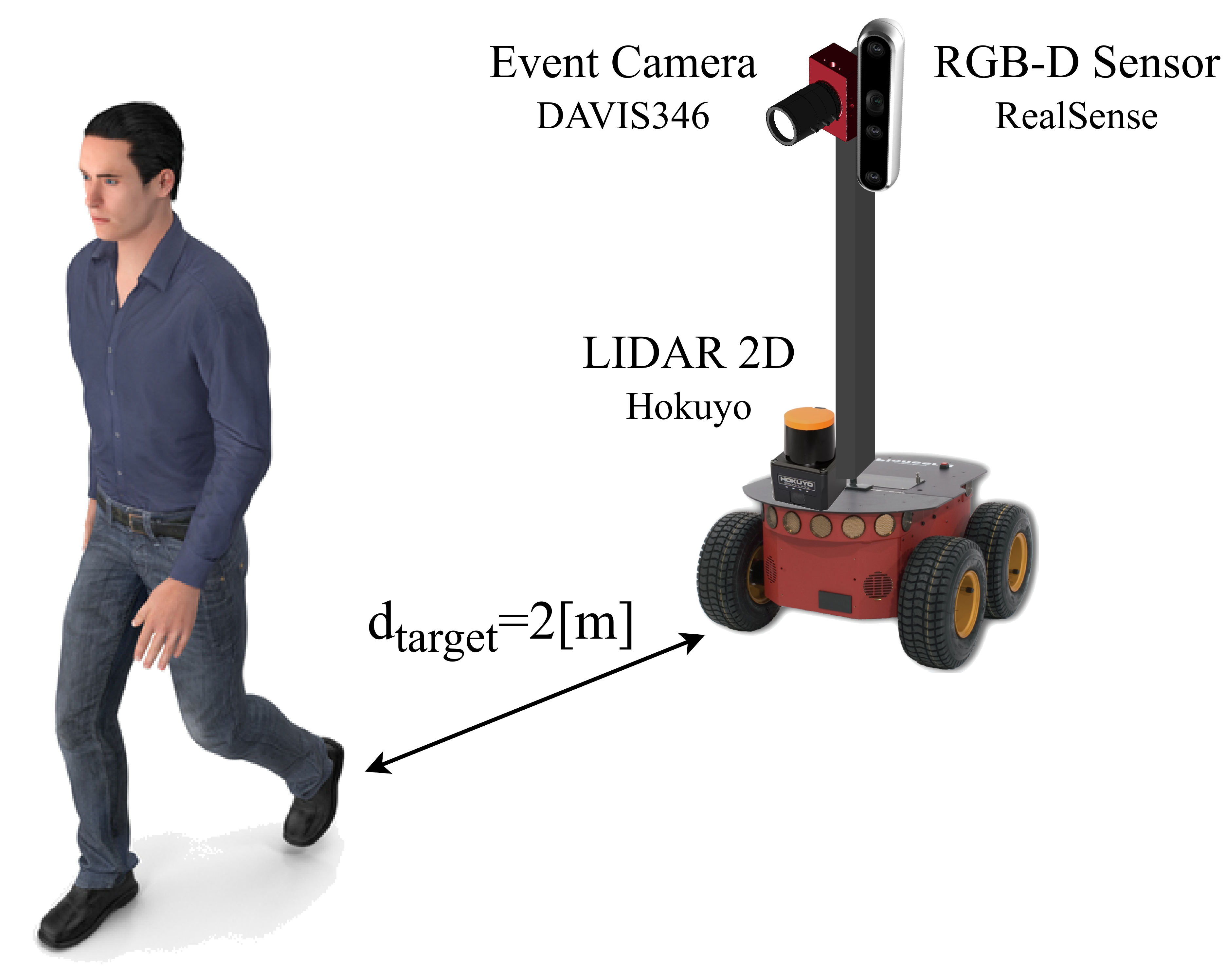}
    \caption{Sensor configuration on the Pioneer 3-AT platform. At the base: Hokuyo 2D (LIDAR). On top: A DAVIS346 (event camera) coupled with a Real Sense (RGB-D) sensor. The target distance between the mobile robotic platform and the pedestrian is 2[m].}
    \label{fig:pioneer3at}
\end{figure}

This work develops a controller for human-following in social robots 
, using a Pioneer 3-AT base equipped with a 2D Hokuyo LIDAR, DAVIS346 event camera, and RealSense camera mounted at 0.8 [m]. The vertically oriented cameras provide full-body pedestrian detection from 1 [m]. The controller prioritizes horizontal position (\(x_{bbox}\)) from the DAVIS346 and distance (\(d_{ped}\)) from the RealSense, discarding vertical variability (\(y_{bbox}\)) for stability. Depth tracking uses a homography to map event coordinates \((x_{events}, y_{events})\) from the DAVIS346 to the RealSense frame for pedestrian distance estimation:
\begin{equation}
\left[ \begin{array}{c} x_{rgbd} \\ y_{rgbd} \end{array} \right] = \text{tf}_{DAVIS \rightarrow RealSense} \left[ \begin{array}{c} x_{events} \\ y_{events} \\ 1 \end{array} \right].
\end{equation}
The LIDAR ensures obstacle avoidance within a 0.5-meter range. Figure \ref{fig:pioneer3at} shows the sensor setup.

\subsection{Simulated scenario design}

\begin{figure}[!h]
\centering
    \subcaptionbox{}
    [.15\textwidth]{\includegraphics[width=\linewidth]{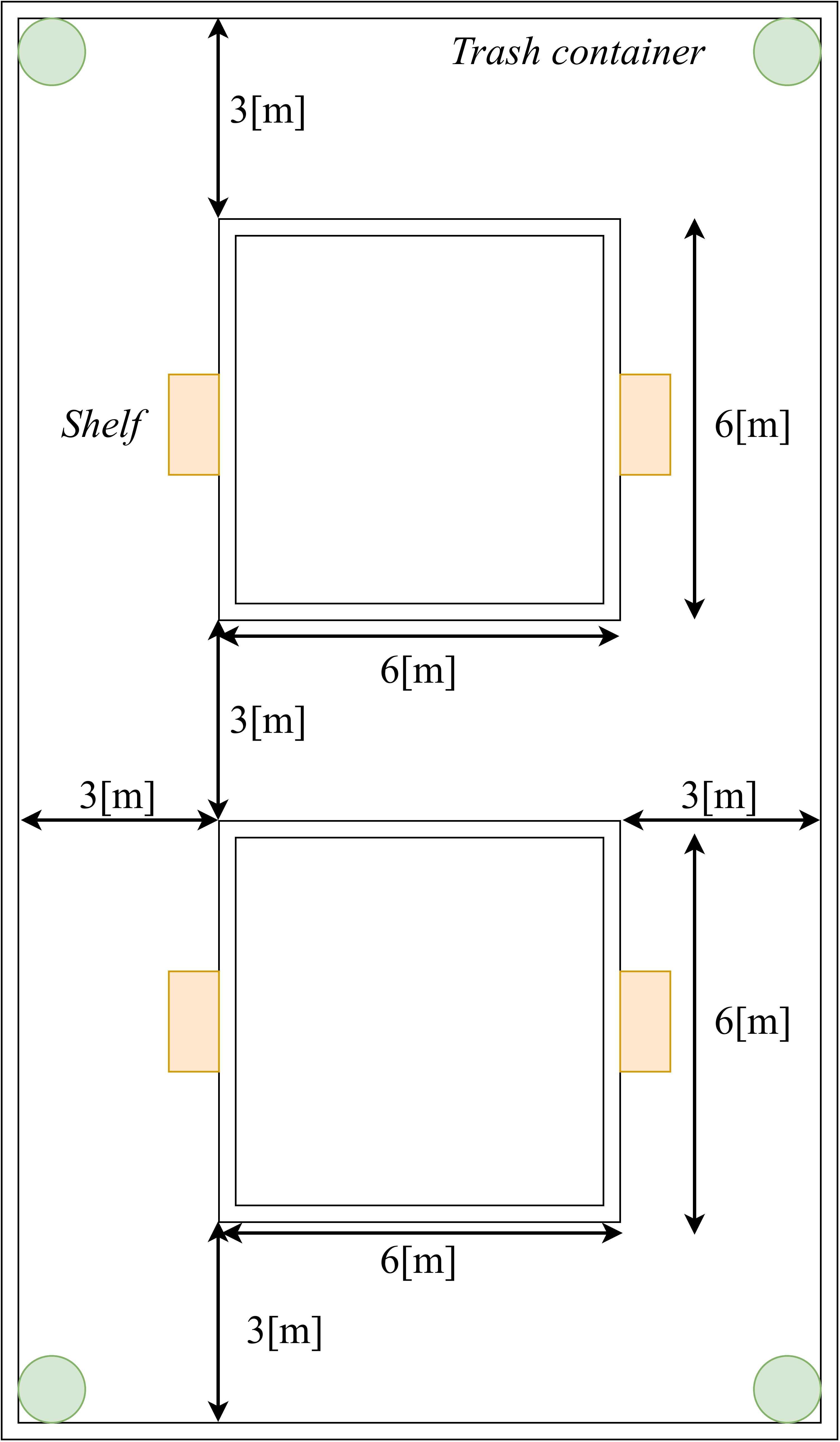 } }   
    \subcaptionbox{}
    [.16\textwidth]{\includegraphics[trim={5cm 0 5cm 0},clip,width=\linewidth]{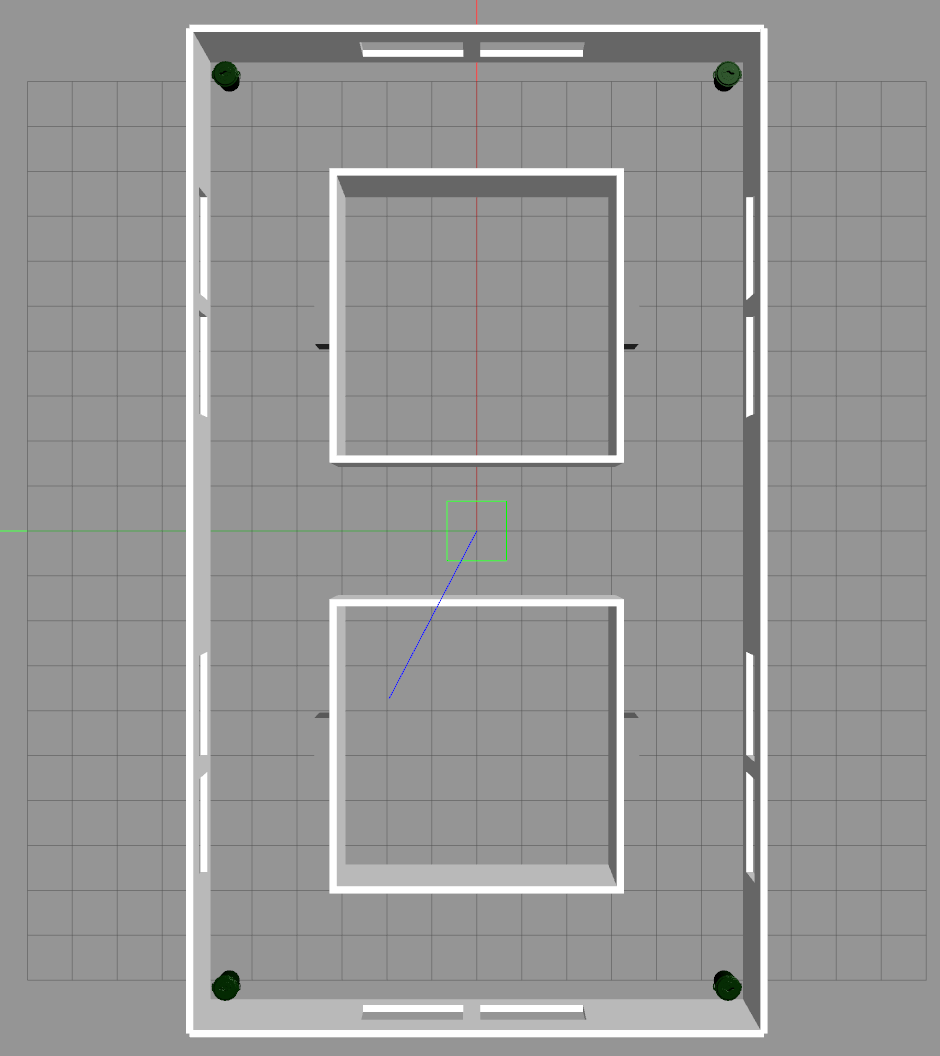} }
    \subcaptionbox{}
    [.15\textwidth]{\includegraphics[width=\linewidth]{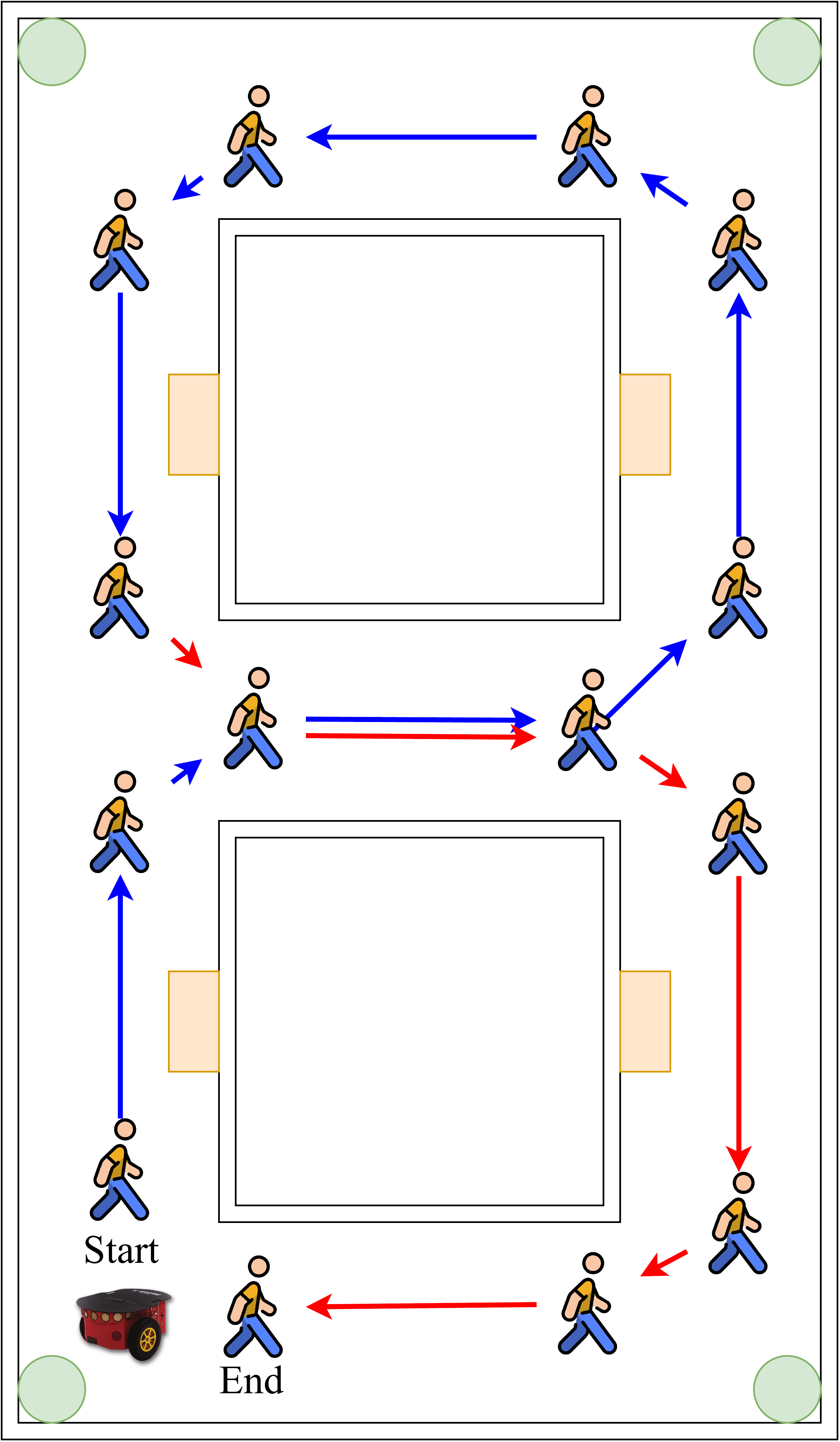} }
    
    \caption{Simulated scenario design: a) Scenario layout in Gazebo; (b) Scenario rendering; (c) Person's trajectory.}
    \label{fig:maps}
\end{figure}

To exploit the high temporal resolution and non-motion blur of event cameras, a 21x12 $[m^2]$ scenario was designed, as shown in Figure \ref{fig:maps}. The map has wide corridors and obstacles, and the pedestrian moves at 0.7 [m/s]. The robot navigates a ``figure-eight''trajectory, combining linear and angular movements over 100 seconds.

\subsection{Using events for real-time navigation controllers}

Event cameras operate asynchronously, eliminating the need for fixed-rate sampling and enabling continuous, low-latency sensing for navigation. This asynchronous nature allows data to be accumulated flexibly for perception, using either fixed-time windows, \( \mathcal{E}_{\Delta t}(t) = \{ e_k \mid t - \Delta t < t_k \leq t \} \), or fixed-event windows, \( \mathcal{E}_N(t) = \{ e_{k-N+1}, \dots, e_k \} \) as studied in \cite{verschae2023event}. These allow the system to adapt temporal resolution and data density according to the dynamics of the scene.

As a result, event data can be accessed and processed over arbitrary intervals, enabling the controller to ``look back'' or ``look ahead'' in time. This supports adaptive inference: the controller can defer computation when activity is low, or update control actions more frequently during fast scene changes—improving responsiveness without increasing computational load.

\subsection{Event-based pedestrian detection}

To address the challenges of high-speed pedestrian detection (such as blur motion), this work proposes training YOLOv5s on the \texttt{PEDRo} database \cite{bbpprsPedro2023}, a specialized event-based dataset designed for person detection in service robotics. 
The data was collected using a moving DAVIS346 event camera in a wide variety of scenarios and lighting conditions and is composed of:
\begin{enumerate*}[label=\roman*)]
\item 119 recordings with an average duration of 18 seconds,
\item 43,259 bounding boxes manually annotated and
\item 27,000 samples.
\end{enumerate*}

\begin{figure}[!h]
\centering
    \includegraphics[width=\linewidth]{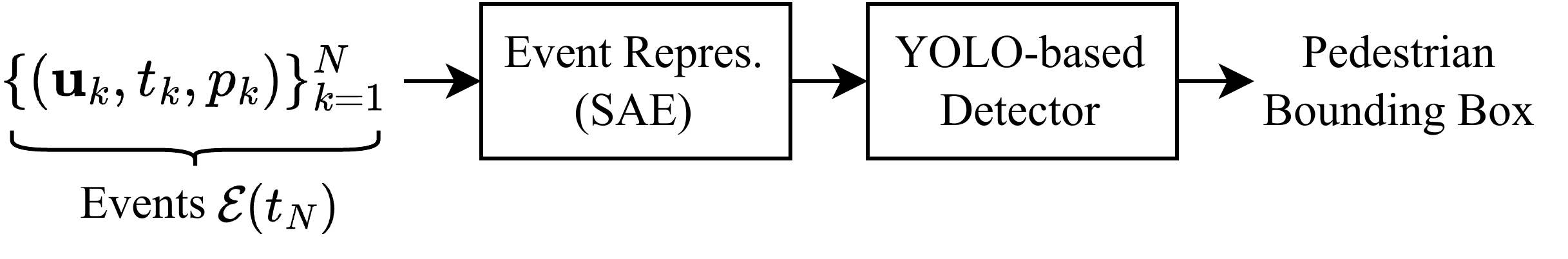}
    \caption{Event-based high-speed pedestrian detection pipeline.}
    \label{fig:event_detector_pipeline}
\end{figure}

Leveraging event cameras, which are significantly less prone to motion blur than conventional frame-based cameras~\cite{v2e}, this work employs the SAE representation \cite{Benosman2014, Lagorce2017-ka} for enhanced detection accuracy, as shown in Figure~\ref{fig:event_detector_pipeline}.

YOLOv5s was selected due to its suitability for real-time applications, as the \texttt{small} variant is optimized for lightweight and fast inference on resource-constrained platforms. Although newer models like YOLOv9 or YOLOv11 may offer slightly better accuracy, state-of-the-art benchmarks \cite{khanam2025yolo} indicate that YOLOv5 achieves significantly faster inference times.

\subsection{Imitation Learning of an PD-based navigation controller}

We initially propose the design of a PD-based controller to calculate the Pioneer 3-AT’s velocities based on the bounding box position \( x_{bbox}, y_{bbox} \) of the person, with \( x_{target} = 173 \), \( y_{target} = 140 \). YOLO handles detection, RealSense provides distance to the pedestrian, and LIDAR ensures obstacle avoidance. 

Now, the few successful expert trajectories generated by this controller are used to train an actor-network (MLP with six inputs and two hidden layers with 30 neurons) using Imitation Learning, with the state \( s \) comprising robot velocities, bounding box data, pedestrian distance, and the LIDAR sample. The 2D output adjusts linear and angular velocities, and training with Mean Squared Error (MSE) approximates expert actions, accelerating RL convergence in DDPG.

\begin{algorithm}[!h]
\caption{Behavior Cloning}
\begin{algorithmic}[1]
\State Obtain a dataset $\mathcal{D}$ containing expert demonstrations
\State Initialize the policy $\pi_{\theta}$ with parameters $\theta$
\State Select an ad-hoc objective function $\mathcal{L}(\theta)$
\State Optimize $\mathcal{L}(\theta)$ using the data contained in $\mathcal{D}$
\end{algorithmic}
\end{algorithm}

The selected objective function \( \mathcal{L}(\theta) \) minimizes the difference between the predicted actions and the expert actions in the dataset \( \mathcal{D} \). It is defined as:
\begin{equation}
    \mathcal{L}(\theta) = \frac{1}{N} \sum_{t=1}^{N} \left\| \pi_{\theta}(s_t) - a_t \right\|^2,    
\end{equation}
where \( N \) is the number of samples in a training batch, \( s_i \) represents the state, and \( a_i \) is the expert's reference action for state \( s_i \). The action predicted by the Actor-network for \( s_i \), \( \pi_{\theta}(s_i) \) corresponds to the action that the model attempts to approximate.

\subsection{Reinforcement Learning for human-robot navigation}

\begin{figure*}[!h]
\centering
    \includegraphics[width=1.0\linewidth]{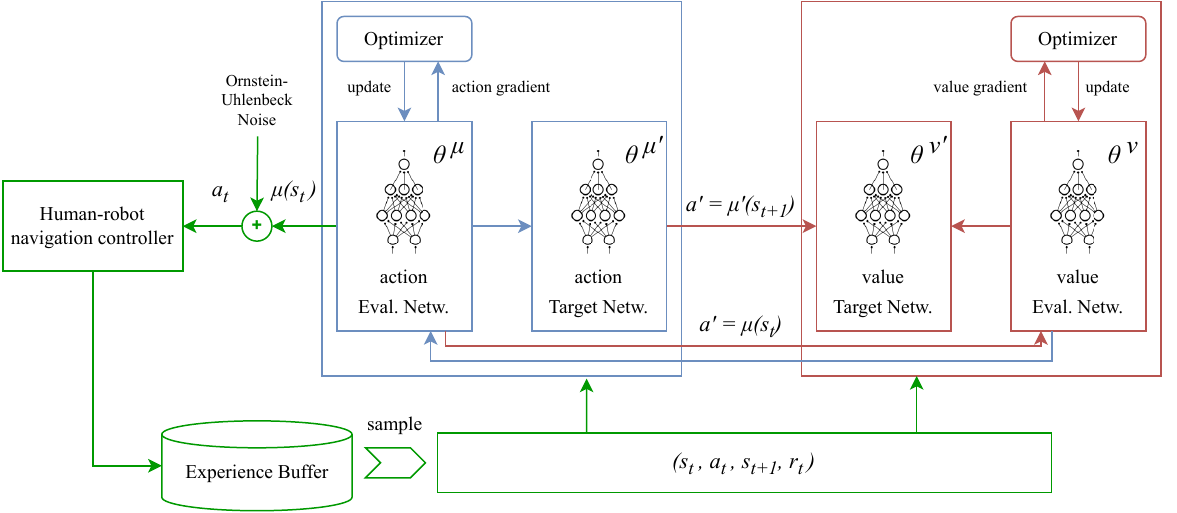}
    \caption{DDPG model-free algorithm applied to estimate velocity increments for robot navigation and human tracking. The state \( s_t = [v_r, w_r, x_{\text{box}}, d_{\text{ped}}, d_{\text{obs}}, \theta_{\text{obs}}] \) includes the robot's linear and angular velocities, the center of the pedestrian’s bounding box, the distance to the person, and the distance \( d_{\text{obs}} \) and angle \( \theta_{\text{obs}} \) to the closest obstacle detected by the LIDAR. Based on \cite{gao2020rl}.}
    \label{fig:ddpg}
\end{figure*}

Reinforcement learning involves an agent learning to map situations to actions to maximize a reward, typically modeled as a Markov Decision Process (MDP). In this work:
\begin{enumerate}
    \item The state \( s_t = [v_r, w_r, x_{\text{box}}, d_{\text{ped}}, d_{\text{obs}}, \theta_{\text{obs}}] \) is a 6-dimensional vector that includes the robot's linear and angular velocities, the center of the pedestrian’s bounding box, the distance to the person, and the distance \( d_{\text{obs}} \) and angle \( \theta_{\text{obs}} \) to the closest obstacle detected by the LIDAR.
    \item Actions \( a_t = [\Delta v_r, \Delta w_r] \) are continuous increments for linear and angular velocities, applied every 0.1 seconds.
    \item The reward function \( r_t \) encourages staying centered on the pedestrian and keeping a target distance of 2 [m]:
    \begin{equation}
    r_t = \begin{cases} 
          500, & \text{if } |e_x| < 1 \text{ pix and } |e_{d_{ped}}| < 0.1 \text{ m}, \\
          -500, & \text{if } d_{\text{obs}} \leq d_{\text{col\_min}}, \\
          -500, & \text{if } d_{ped} > 3 \text{ m or } d_{ped} < 1 \text{ m}, \\
          r_{t_{eq}} & \text{otherwise},
       \end{cases}
    \end{equation}
    where \( r_{t_{eq}} \) penalizes deviations in centering and distance, defined as
    \begin{equation}
        r_{t_{eq}} = -k_{\text{dist}}  |e_{d_{\text{ped}}}|^2 - \left( \frac{k_x}{1 + \alpha |e_{d_{\text{ped}}}|^2} \right)  |e_x|^2,
    \end{equation}
    with a factor $\alpha$ to adjust the influence of the centering error \( e_x \) when the distance error \( e_{d_{\text{ped}}} \) is significant.
\end{enumerate}

\subsection{DDPG for Optimal Policy in RL}

DDPG is a model-free, off-policy reinforcement learning algorithm designed for continuous action spaces well-suited for optimizing control policies like linear and angular velocities. In DDPG, actor-critic networks are used, where the actor-network \( \mu(s|\theta^\mu) \) proposes actions, and the critic network \( Q(s, a|\theta^Q) \) evaluates them. To stabilize training, each network has a corresponding target network \( \mu'(s|\theta^{\mu'}) \) and \( Q'(s, a|\theta^{Q'}) \), updated incrementally to ensure slow changes that promote convergence:
\begin{equation}
    \theta^{\mu'} \leftarrow \tau \theta^\mu + (1 - \tau) \theta^{\mu'}, \quad \theta^{Q'} \leftarrow \tau \theta^Q + (1 - \tau) \theta^{Q'},    
\end{equation}
where \( \tau \) is the update rate.
The critic network is trained by minimizing the temporal difference loss:
\begin{equation}
    L\left(\theta^Q\right) = \mathbb{E}_{\beta, p} \left[ \left( Q \left( s_t, a_t | \theta^Q \right) - y_{i,t} \right)^2 \right],
\end{equation}
where the target value \( y_{i,t} \) is calculated as:
\begin{equation}
    y_{i,t} = r\left( s_t, a_t \right) + \gamma Q' \left( s_{t+1}, \mu' \left( s_{t+1} | \theta^{\mu'} \right) | \theta^{Q'} \right).
\end{equation}
Meanwhile, the actor network is updated by following the gradient:
\begin{equation}
    \label{eq:actor_network_updating}
    \nabla_{\theta^{\mu}} J = \mathbb{E}_{s_t} \left[ \nabla_{a} Q \left( s, a | \theta^{Q} \right) |_{a = \mu(s|\theta^{\mu})} \nabla_{\theta^{\mu}} \mu(s | \theta^{\mu}) \right].
\end{equation}
The target networks are used to promote the convergence of the DDPG algorithm and serve as delayed copies of the original networks. The parameters of the target networks are updated as follows:
\begin{equation}
    \label{eq:target_network_updating}
    \theta^{\mu'} = \tau \theta^{\mu} + (1 - \tau) \theta^{\mu'}, \quad \theta^{Q'} = \tau \theta^{Q} + (1 - \tau) \theta^{Q'},
\end{equation}
where \( \tau \) determines the update rate. 
A replay buffer is used to store and sample experience tuples \( (s_t, a_t, r_t, s_{t+1}) \) for uncorrelated training, enhancing learning stability. To encourage effective exploration in continuous spaces, Ornstein-Uhlenbeck noise is added to actions, providing temporally correlated noise.
The DDPG process is summarized in Figure \ref{fig:ddpg} and Algorithm \ref{alg:ddpg}, illustrating its application to velocity control in human-robot navigation.
\begin{algorithm}[!h]
\caption{DDPG for human-robot navigation controller.}
\label{alg:ddpg}
\begin{algorithmic}[1]
\State Initialize $\theta^{\mu}$ with pre-trained weights from I.L.
\State Initialize $\theta^{Q}$ randomly
\State Initialize target networks by copying the original networks: $\theta^{\mu'} \gets \theta^{\mu}, \theta^{Q'} \gets \theta^{Q}$
\State Initialize a noise generator $\Omega$
\For{episode $= 1, \dots, n$}
    \State Randomly initialize the robot's position and observe the initial state $s_1$
    \For{$t = 1, \dots,$}
        \State Observe the state $s_t$ of the environment
        \State Estimate the linear and angular velocity increments $\Delta v_r, \Delta w_r = \beta(s_t) = \mu(s_t|\theta^{\mu}) + \Omega_t$
        \State Compute the new linear and angular velocities
        \State Observe the reward $r_t$ and the next state $s_{t+1}$
        \State Store the tuple $(s_t, a_t, r_t, s_{t+1})$ in the replay buffer $M$
        \If{$s_{t+1}$ is a terminal state}
            \State \textbf{break}
        \EndIf
        \If{$M$ exceeds a threshold}
            \State Randomly sample $N$ tuples $(s_t, a_t, r_t, s_{t+1})$ from buffer $M$ to train the critic and actor networks according to Eq. \ref{eq:actor_network_updating}
            \State Update the parameters of the target networks according to Eq. \ref{eq:target_network_updating}
        \EndIf
    \EndFor
\EndFor
\end{algorithmic}
\end{algorithm}

\subsection{Training details}

The simulated scenario uses Gazebo and ROS to conduct the training process. The models are implemented in Python, and the neural networks are built using the \textit{PyTorch} framework. The episode will terminate and restart once these term states are found. In this work, the term states include the following:

\begin{enumerate}
    \item The human reaches the final position.
    \item Lose the feature in the camera's field of view for more than 4 seconds.
    \item The estimated distance between the camera and the person is greater than 3 [m] or less than 1 [m].
    \item There is an obstacle at a distance of less than 0.5 [m].
\end{enumerate}

Additionally, all DDPG neural networks have two hidden layers. The units for the first and second hidden layers are 30 and 30, respectively. The actor and target networks take the state as input and the 2-D action \( a_1, a_2 \) as output. The critical and target networks take the state-action pair 6-D \((s, a)\) as input and the action value function 2-D \( Q(s, a) \) as output. The activation functions used in the hidden layers of these neural networks are \texttt{ReLU} functions. In addition, modified \texttt{tanh} activation functions are added in the output layers of the actor and actor-target networks to ensure that their outputs are in the range \([-0.2, 0.2]\) for $\Delta v_r$ and \([-0.5, 0.5]\) for $\Delta w_r$.
The other parameters are reported in Table \ref{tab:parameters_ddpg}.

\begin{table}[!h]
\centering
\small
\caption{Experimental parameters and network architecture for the DDPG agent.}
\begin{tabular}{|l|c|}
\hline
\textbf{Parameter} & \textbf{Value} \\ \hline
Discount factor & 0.99 \\ \hline
Buffer size & 1,000,000 \\ \hline
Batch size & 64 \\ \hline
Actor learning rate & $1 \times 10^{-4}$ \\ \hline
Critic's learning rate & $1 \times 10^{-4}$ \\ \hline
$\tau$ (to update target networks) & 0.001 \\ \hline
O.U. Noise for $\Delta v$ (mu, theta, sigma) & (0.0, 0.0, 0.2) \\ \hline
O.U. Noise for $\Delta w$ (mu, theta, sigma) & (0.0, 0.2, 0.3) \\ \hline
Linear speed range & [0, 1] [m/s] \\ \hline
Angular speed range & [-0.5, 0.5] [rad/s] \\ \hline
Actor network (MLP) & 6--30--30--2 \\ \hline
Critic network (MLP) & (6+2)--30--30--1 \\ \hline
\end{tabular}
\label{tab:parameters_ddpg}
\end{table}

%% file: 05_results.tex
\section{Results}
\label{sec:results}

This section presents the results achieved after 
\begin{enumerate*}[label=\arabic*)]
\item training an event-based pedestrian detector with the PEDRo database, 
\item and training an RL-based robotic platform controller using Behavioral Cloning to reduce the agent's initial policy learning time.
\end{enumerate*}

\subsection{Pedestrian detection based on event cameras}

\begin{figure}[!h]
\centering
    \subcaptionbox{Learning curves of the YOLO-based pedestrian detector.}
    [.48\textwidth]{\includegraphics[width=\linewidth]{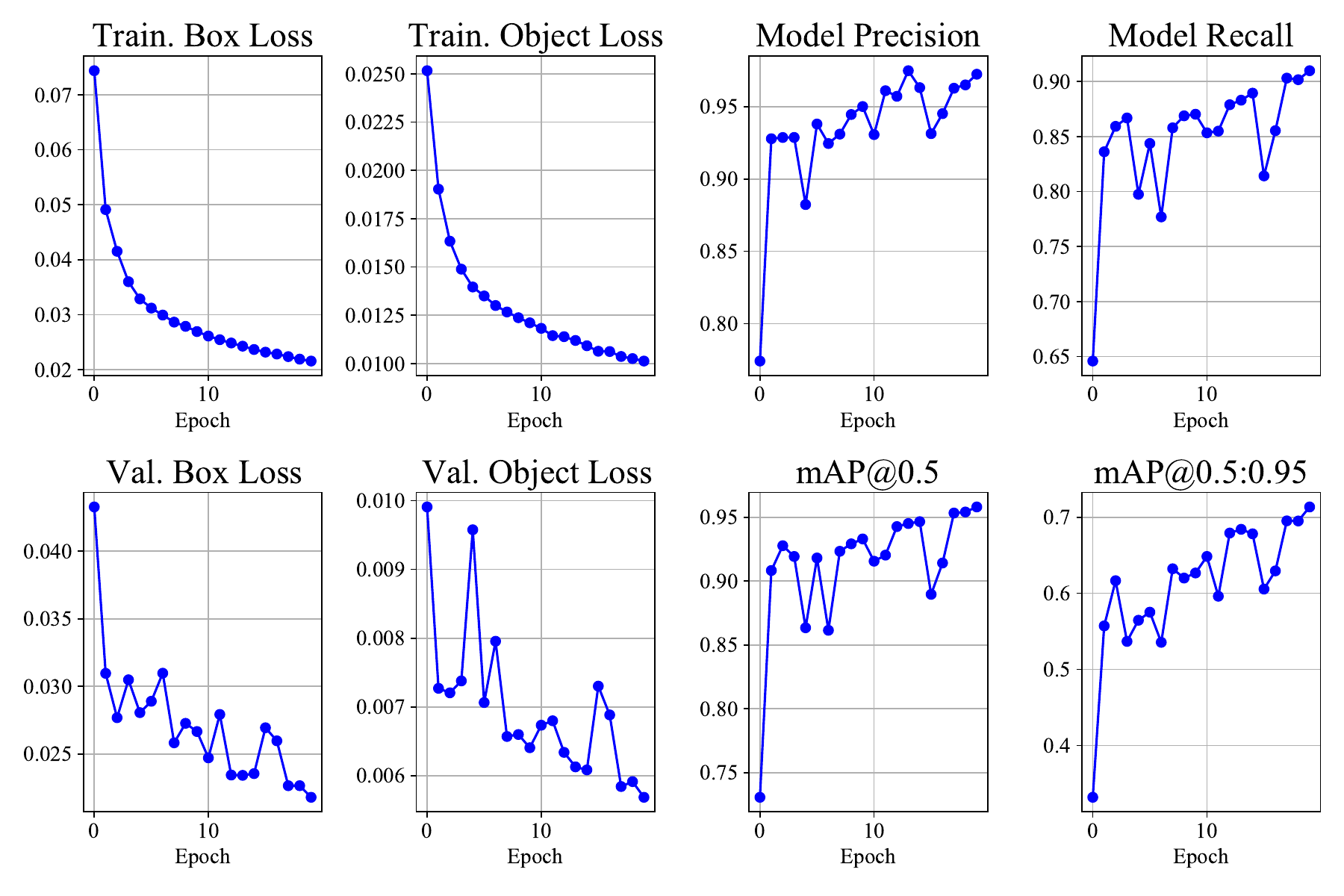} }
    \hfill
    \subcaptionbox{Visual samples of the YOLO-based pedestrian detector.}
    [.48\textwidth]{\includegraphics[trim={0 0 0 580},clip,width=\linewidth]{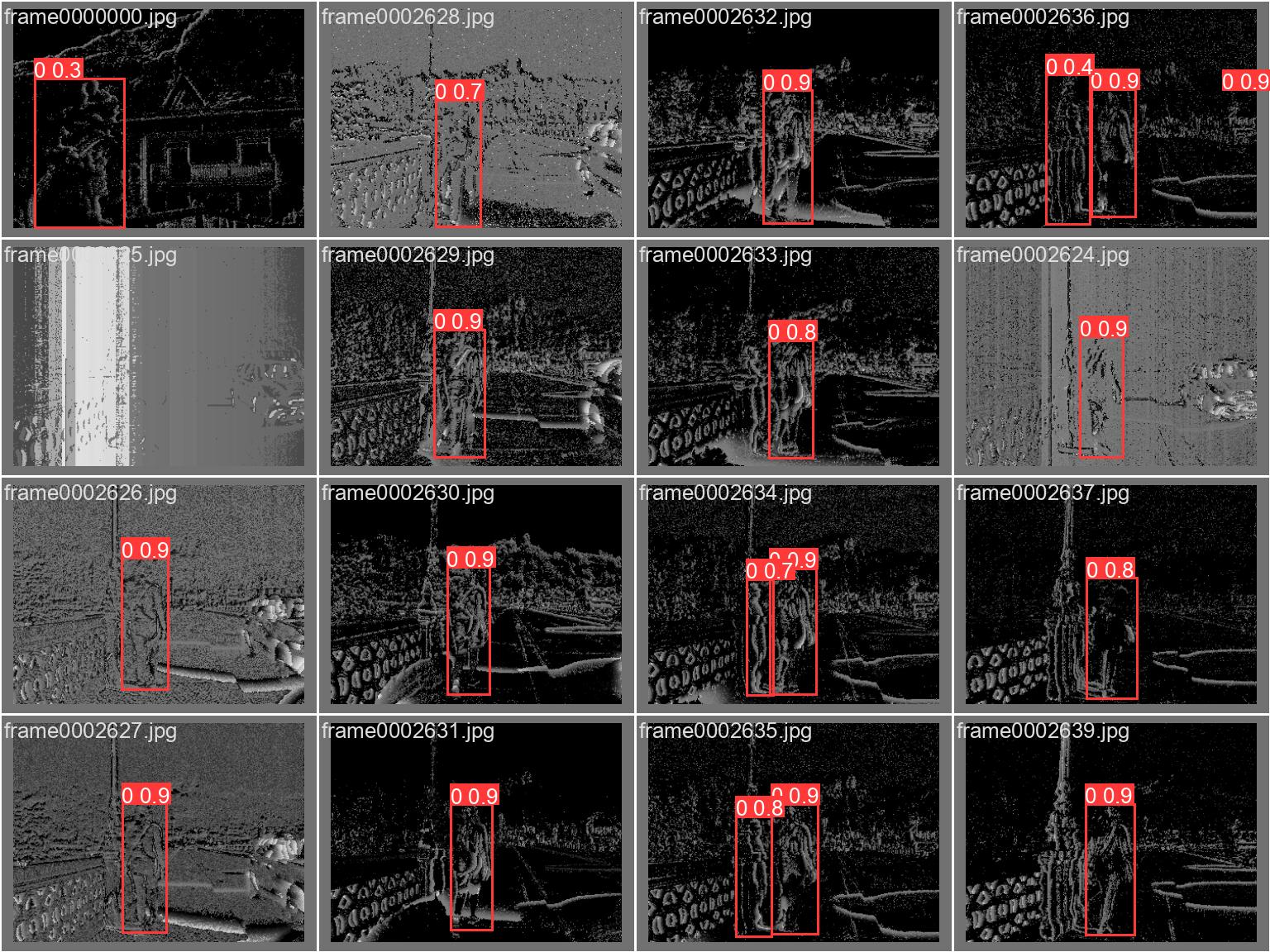} }
    
    \caption{Results after training a YOLOv5s detector on the \texttt{PEDRo} database \cite{bbpprsPedro2023} using the SAE event representation with 10$[ms]$ time window.}
    \label{fig:event-based_ped_performance}
\end{figure}

Figure \ref{fig:event-based_ped_performance} reports the results obtained after training the YOLO network on the PEDRo event-based database for high-speed pedestrian detection.

\begin{table}[!h]
    \small
    \centering
    \captionsetup[subtable]{skip=0pt}
    \caption{Performance statistics for human-robot navigation controllers: $v_r$ linear velocity, $\omega_r$ angular velocity, $x_{\text{box}}$ bounding box x-center value, $d_{\text{ped}}$ distance to pedestrian.}
    \label{tab:combined-controllers}
    \begin{subtable}[t]{\linewidth}
    \centering
    \caption{PD-based Controller}
    \label{tab:pd-controller}
    \begin{tabular}{l|c|c|c|c|}
      \cline{2-5}
      & \textbf{$v_r$ [m/s]} & \textbf{$\omega_r$ [m/s]} & \textbf{$x_{\text{box}}$ [px]} & \textbf{$d_{\text{ped}}$ [m]} \\ \hline
      \multicolumn{1}{|l|}{\textbf{Mean}}      & 0.6923  & 0.0328  & 167.0698  & 2.5889  \\ \hline
      \multicolumn{1}{|l|}{\textbf{Median}}    & 0.6923  & 0.0318  & 167.0000  & 2.6414  \\ \hline
      \multicolumn{1}{|l|}{\textbf{Std. Dev.}} & 0.1782  & 0.2711  & 46.9139   & 0.2587  \\ \hline
      \multicolumn{1}{|l|}{\textbf{Min.}}      & 0.0000  & -0.5000 & 83.0000   & 1.9580  \\ \hline
      \multicolumn{1}{|l|}{\textbf{Max.}}      & 1.0000  & 0.5000  & 260.0000  & 3.1453  \\ \hline
    \end{tabular}
    \end{subtable}
    \vfill
    \begin{subtable}[t]{\linewidth}
    \centering
    \caption{IL-based Controller}
    \label{tab:il-controller}
    \begin{tabular}{l|c|c|c|c|}
      \cline{2-5}
      & \textbf{$v_r$ [m/s]} & \textbf{$\omega_r$ [m/s]} & \textbf{$x_{\text{box}}$ [px]} & \textbf{$d_{\text{ped}}$ [m]} \\ \hline
      \multicolumn{1}{|l|}{\textbf{Mean}}      & 0.6704  & 0.0349  & 172.2215  & 2.0846  \\ \hline
      \multicolumn{1}{|l|}{\textbf{Median}}    & 0.6977  & 0.0093  & 172.0000  & 2.0737  \\ \hline
      \multicolumn{1}{|l|}{\textbf{Std. Dev.}} & 0.1778  & 0.3795  & 34.3104   & 0.1853  \\ \hline
      \multicolumn{1}{|l|}{\textbf{Min.}}      & 0.0740  & -0.5000 & 96.0000   & 1.6653  \\ \hline
      \multicolumn{1}{|l|}{\textbf{Max.}}      & 1.0000  & 0.5000  & 255.0000  & 2.6043  \\ \hline
    \end{tabular}
    \end{subtable}
    
    \begin{subtable}[t]{\linewidth}
    \centering
    \caption{RL-based Controller}
    \label{tab:rl-controller}
    \begin{tabular}{l|c|c|c|c|}
      \cline{2-5}
      & \textbf{$v_r$ [m/s]} & \textbf{$\omega_r$ [m/s]} & \textbf{$x_{\text{box}}$ [px]} & \textbf{$d_{\text{ped}}$ [m]} \\ \hline
      \multicolumn{1}{|l|}{\textbf{Mean}}      & 0.6568  & 0.0376  & 171.2376  & 2.0756  \\ \hline
      \multicolumn{1}{|l|}{\textbf{Median}}    & 0.6692  & 0.0279  & 172.0000  & 2.0696  \\ \hline
      \multicolumn{1}{|l|}{\textbf{Std. Dev.}} & 0.2407  & 0.3603  & 27.4392   & 0.1982  \\ \hline
      \multicolumn{1}{|l|}{\textbf{Min.}}      & 0.0000  & -0.5000 & 112.0000  & 1.6754  \\ \hline
      \multicolumn{1}{|l|}{\textbf{Max.}}      & 1.0000  & 0.5000  & 253.0000  & 2.7543  \\ \hline
    \end{tabular}
    \end{subtable}
    \label{tab:controllers-performance}
\end{table}

\begin{figure*}[t]
\centering
    \subcaptionbox{PD-based controller}
    [.33\textwidth]{\includegraphics[width=\linewidth]{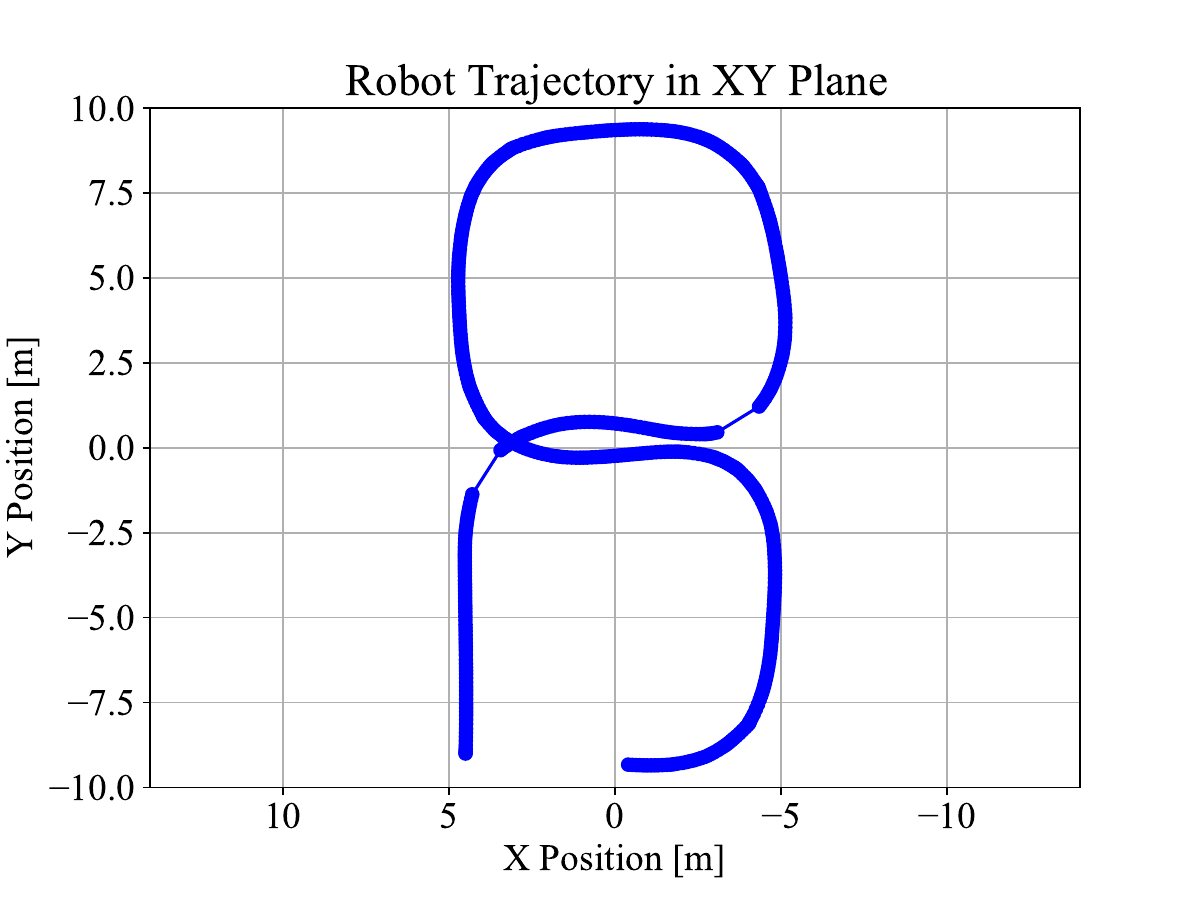 } }    
    \subcaptionbox{IL-based controller}
    [.33\textwidth]{\includegraphics[width=\linewidth]{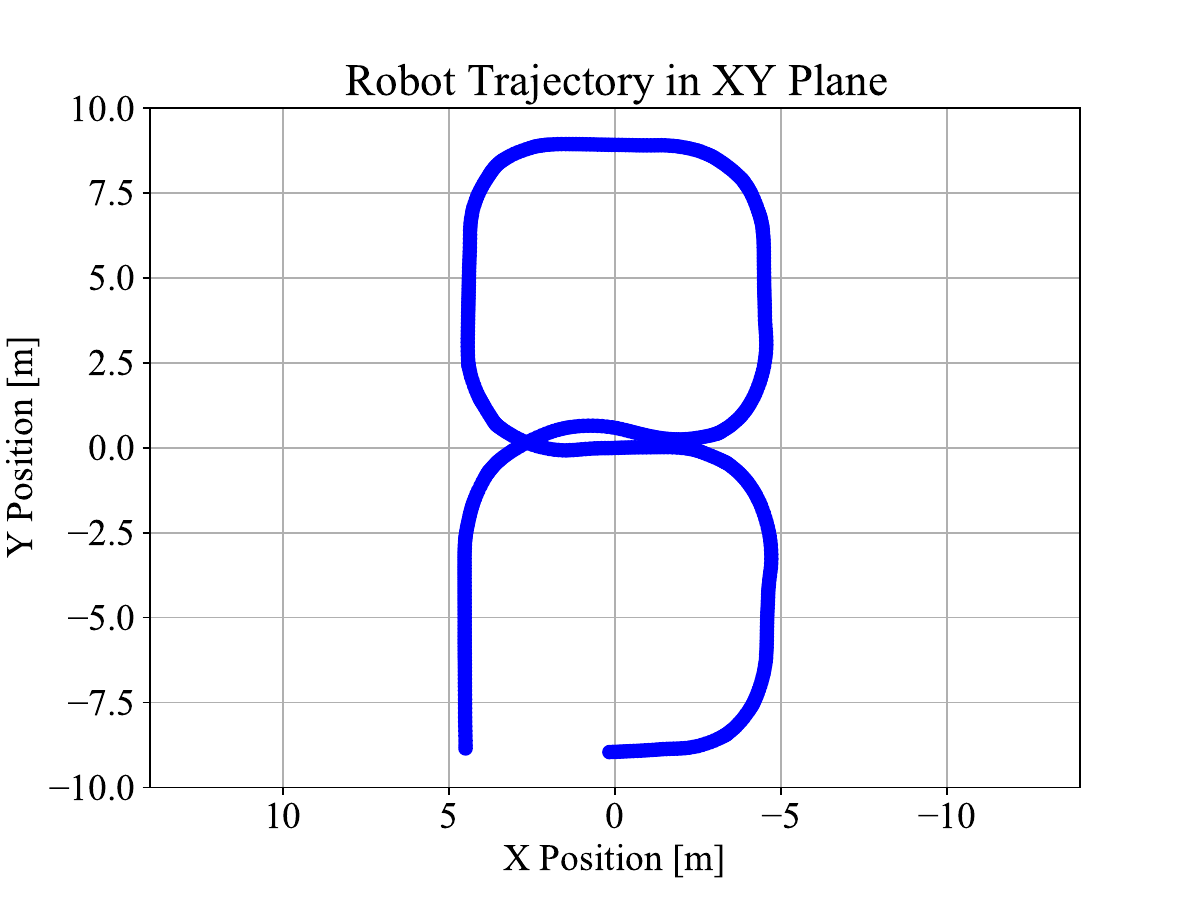 } }
    \subcaptionbox{RL-based controller}
    [.33\textwidth]{\includegraphics[width=\linewidth]{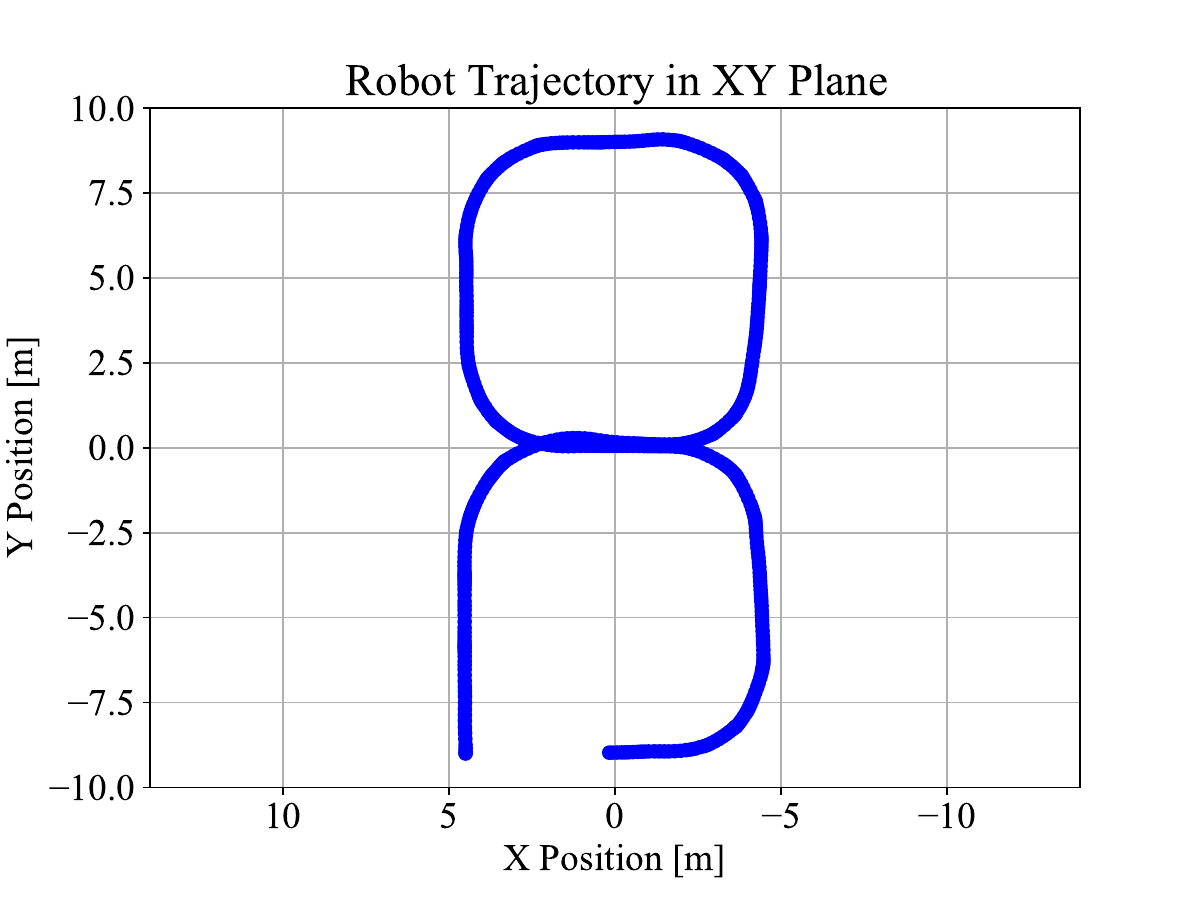 } }

    \caption{Trajectories described after applying different navigation and human-following controllers using event cameras and other relevant sensors: (a) PD-based controller; (b) Imitation Learning-based controller; (c) Reinforcement Learning-based controller.}
    \label{fig:results-1}
\end{figure*}

\begin{figure*}[t]
\centering
    \subcaptionbox{PD-based controller\label{fig:sub:pd}}
    [.33\textwidth]{\includegraphics[width=\linewidth]{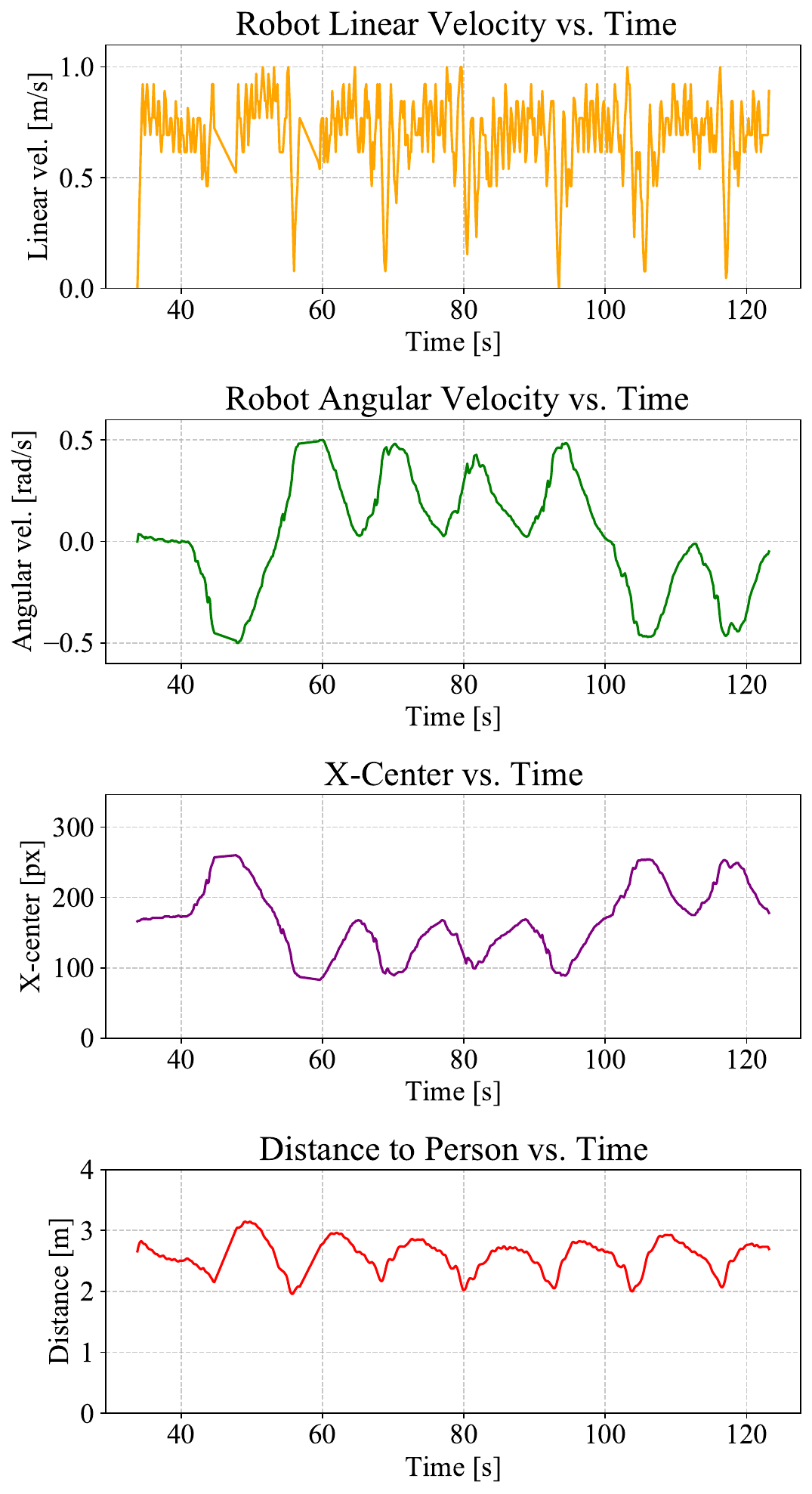 } }
    \subcaptionbox{IL-based controller\label{fig:sub:il}}
    [.33\textwidth]{\includegraphics[width=\linewidth]{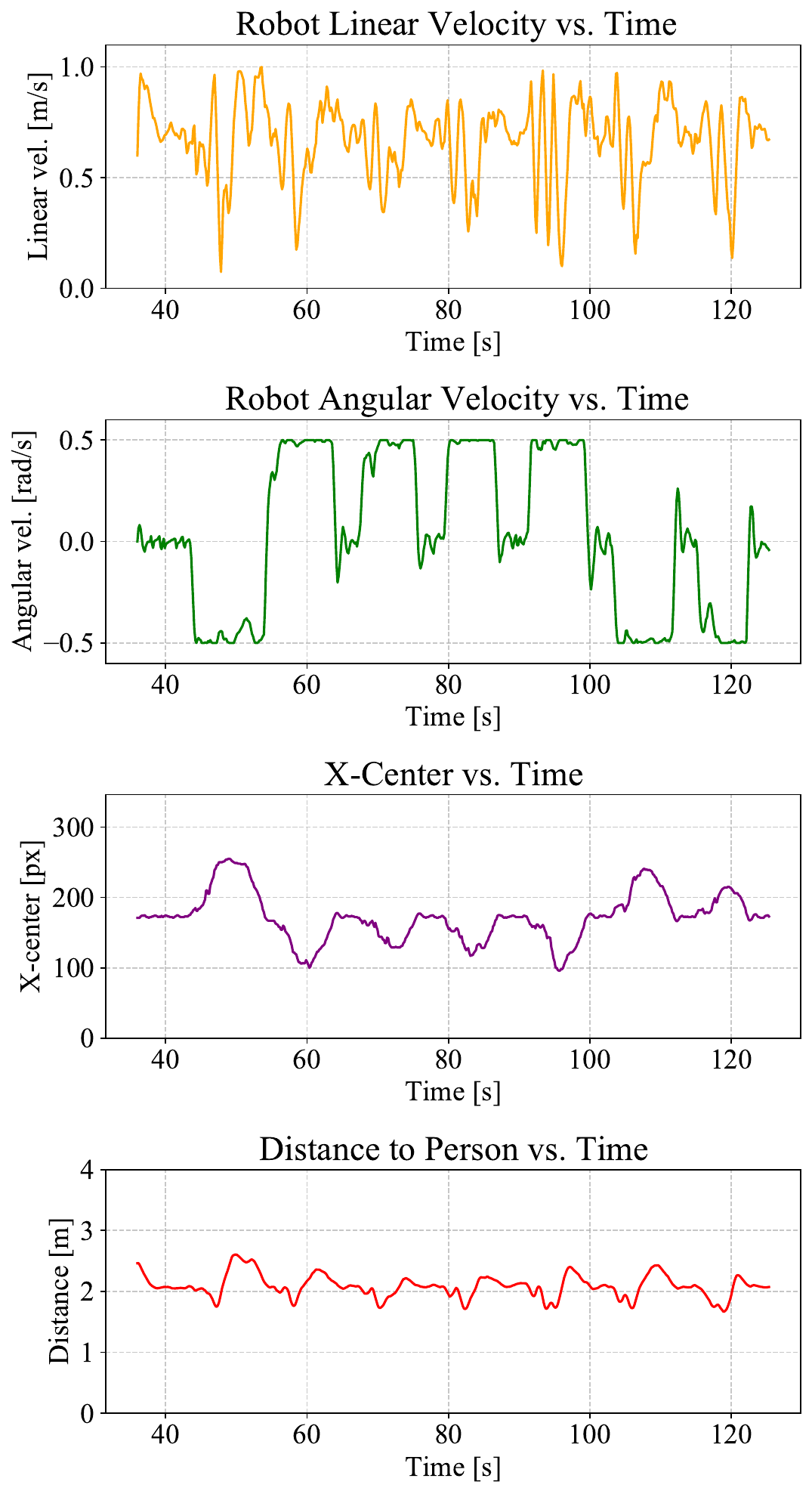 } }
    \subcaptionbox{RL-based controller\label{fig:sub:rl}}
    [.33\textwidth]{\includegraphics[width=\linewidth]{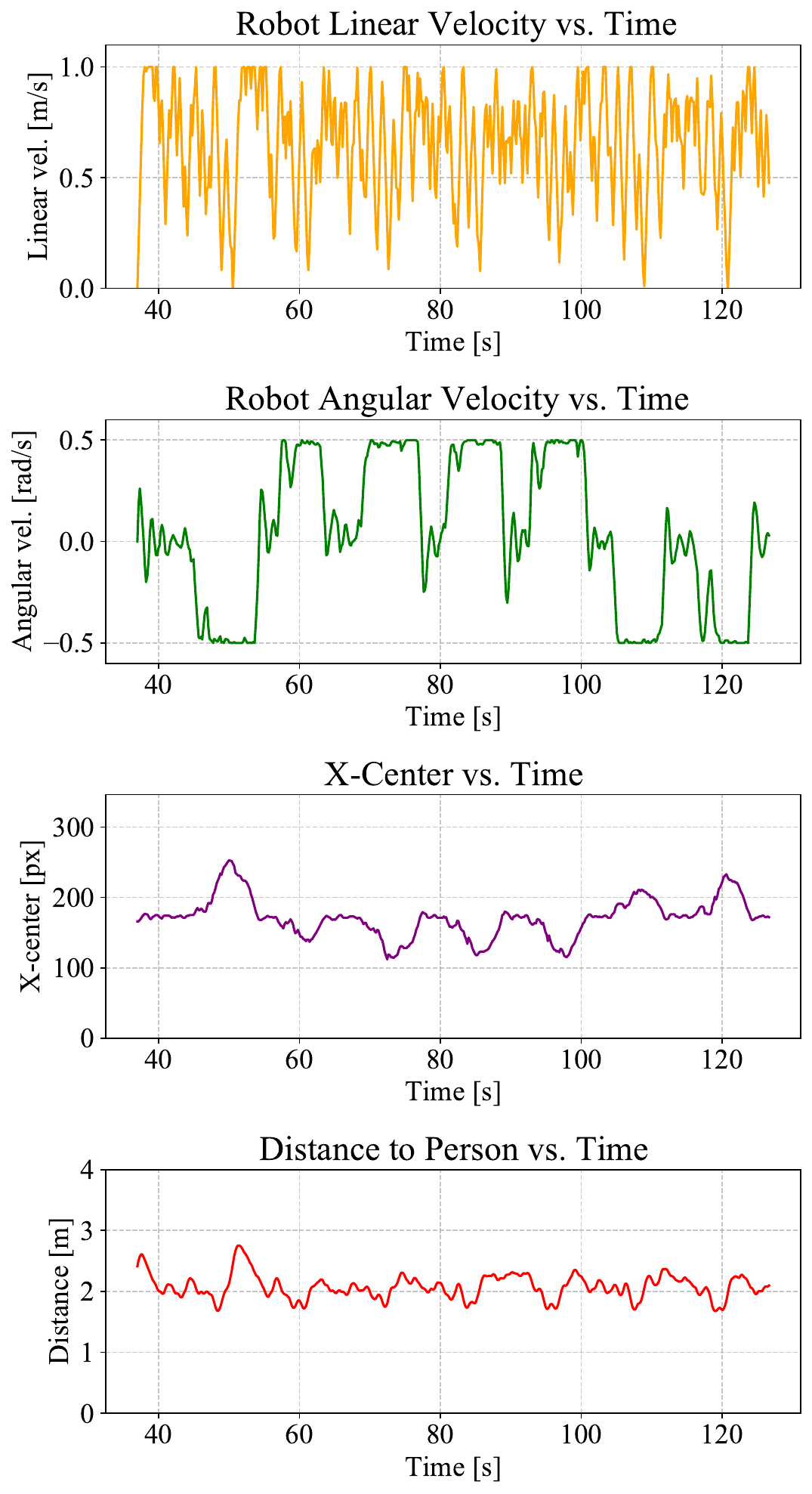 } }
    \caption{Relevant metrics (robot linear/angular velocity, x-y bounding box estimation, distance-to-person) on navigation and human-following controllers using event cameras and other relevant sensors: (a) PD-based controller; (b) Imitation Learning-based controller; (c) Reinforcement Learning-based controller.}
    \label{fig:results-2}
\end{figure*}

Figure \ref{fig:event-based_ped_performance}(a) shows that the loss values for training and validation decrease steadily over epochs, implying that the model is learning to better adjust its predictions for the bounding box and the classification of detected pedestrians.
Accuracy and recall metrics show consistent increases, reaching values close to 0.90, offering good performance in detecting and localizing pedestrians on the scene. In addition, the values of $mAP_{0.5}$ and $mAP_{0.5:0.95}$ also increase throughout the training, improving average accuracy for different Intersection over Union thresholds, reaching values above 0.7 in the most demanding mAP (0.5:0.95).
On the other hand, the results prove that the use of the SAE event data representation is adequate to detect people with a high level of confidence (0.7-0.9) in pedestrian following, using a 10[ms] time window to generate the representation.

\subsection{Evaluation of the navigation algorithms}

\subsubsection{About the described trajectories}

Figure \ref{fig:results-1} compares the simulated XY trajectories of the Pioneer 3-AT using three control strategies for human-following navigation. 
The PD-based controller (Fig. \ref{fig:results-1} (a)) displays considerable deviations, especially in curves, indicating its limited ability to adapt to rapid changes in direction. 
The IL controller (Fig. \ref{fig:results-1} (b)) demonstrates improved stability and follows the target path more closely, though it still shows minor deviations in complex areas, particularly tight curves. This improvement highlights IL’s advantage in achieving greater path adherence compared to the PD-based approach, though it has some limitations.
The RL controller (Fig. \ref{fig:results-1} (c)) produces the most accurate trajectory, showing smooth adaptations and effectively managing changes in orientation. This approach achieves precise path-following and a high-speed response to scene conditions.

\subsubsection{About the complementary metrics}

Table~\ref{tab:controllers-performance} and Figure~\ref{fig:results-2} present the performance metrics for the three control strategies. The analysis considers robot velocity profiles, target centering (\(x_{\text{box}}\)), and distance to the pedestrian (\(d_{\text{ped}}\)).

For the PD-based controller (Table~\ref{tab:pd-controller}, Figure~\ref{fig:sub:pd}), both linear and angular velocities exhibit high variability, with frequent and abrupt changes. This indicates a reactive behavior, constantly correcting to maintain distance and centering. The X-center fluctuates around a mean of 167~[px], deviating from the true target of 173~[px], while the distance to the person varies between 2 and 3 meters, indicating limited control accuracy.

The Imitation Learning-based controller (Table~\ref{tab:il-controller}, Figure~\ref{fig:sub:il}) reduces these oscillations moderately. Although still reactive, it shows improved stability in linear motion and smoother angular corrections. The X-center aligns more closely to the 173~[px] target (mean of 172~[px]), and the distance remains consistently closer to 2~[m] than the PD baseline, improving both lateral and longitudinal tracking.

Finally, the Reinforcement Learning-based controller (Table~\ref{tab:rl-controller}, Figure~\ref{fig:sub:rl}) achieves the best overall performance, but not without some trade-offs. While its linear velocity presents a higher standard deviation (SD = 0.24) compared to the PD and IL controllers, the temporal evolution is smoother and more context-driven. Angular velocity shows consistent and controlled adaptation. The X-center stabilizes near 171~[px], closest to the true target of 173~[px], with reduced variance. Similarly, the distance to the pedestrian is well maintained near the 2~[m] goal. These results highlight the RL controller’s ability to adaptively balance precision and stability across different navigation scenarios.
\begin{figure}[!h]
\centering
    \includegraphics[trim={3cm 1cm 1.5cm 1.5cm},clip,width=\linewidth]{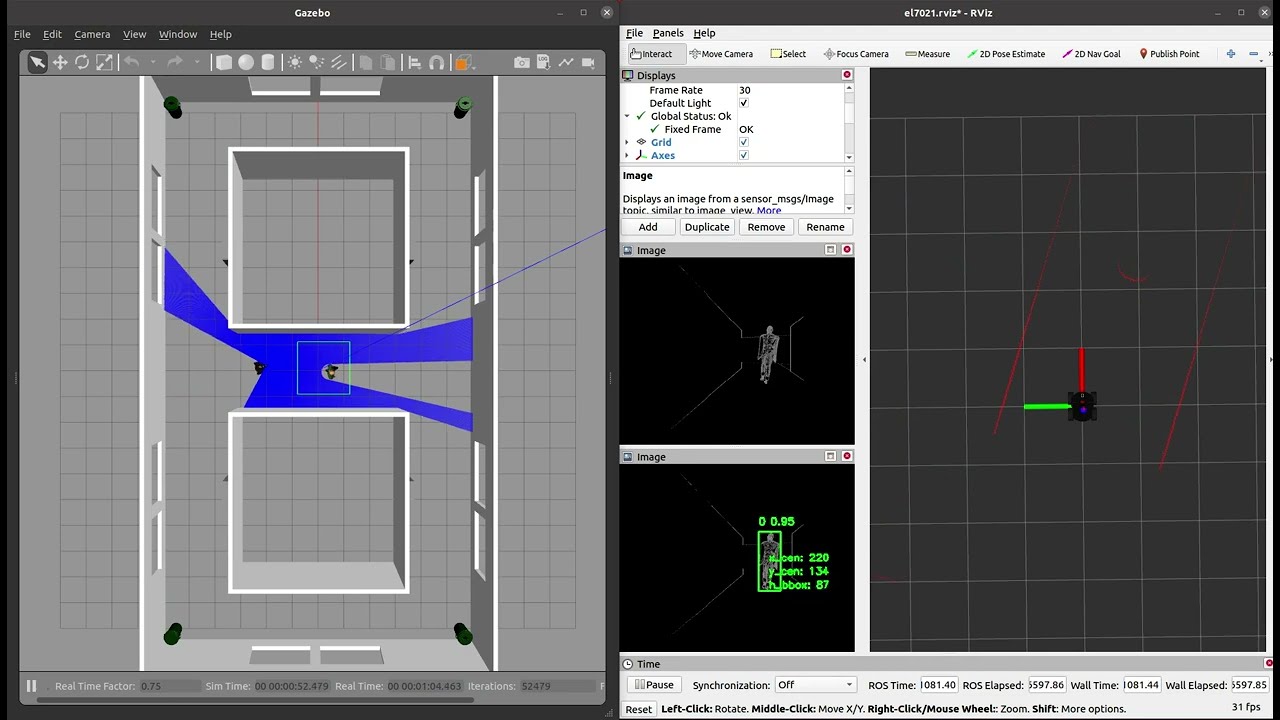 }
    \caption{Visual example of the RL-based controller simulated performance for human following. On the right, the figure shows the map rendered in Gazebo. On the left, the RViz software shows the event-based pedestrian detection using SAE and a YOLOv5s detector, the obstacles using a Hokuyo2D, and the Pioneer3AT.}
    \label{fig:rl_gazebo_example}
\end{figure}

%% file: 10_conclusion.tex
\section{Conclusions and Future Work}
\label{sec:conclusions}

This work presents the design and implementation of a reinforcement learning-based controller for human-robot navigation using event cameras, depth, and LIDAR data. The proposed system outperforms traditional PD and IL-based approaches in simulation, offering better centering, distance regulation, and adaptation to trajectory changes. These results highlight the potential of event-based sensing for human-centered robotic navigation.

However, the results are not yet definitive. As future work, we aim to develop a complete end-to-end system fully based on event-based perception, capable of operating in real-world conditions. This includes testing in diverse pedestrian motion patterns, extending the evaluation to more complex environments, exploring alternative RL algorithms (e.g., TD3), and studying different rewards. Additional efforts should focus on exploring alternative event-based representations and developing monocular event-based methods for depth and pose estimation to reduce the reliance on multiple sensors.

\section*{Acknowledgment}

This work was partially supported by the FONDEQUIP project EQM17004 and the Basal project AFB230001. 
A special mention to Ariel Zuniga (Universidad de O'Higgins) for his support with ROS.